\documentclass[10pt,twocolumn,letterpaper]{article}

\usepackage{eso-pic}
\usepackage{iccv}
\usepackage{times}
\usepackage{epsfig}
\usepackage{graphicx}
\usepackage{amsmath}
\usepackage{amssymb}
\usepackage{times}
\usepackage{graphicx}
\usepackage{kotex}
\usepackage{kotex-logo}
\usepackage[linesnumbered,ruled]{algorithm2e}
\usepackage{caption}
\usepackage{subcaption}
\usepackage{multirow}
\usepackage{url,gensymb}
\newsavebox{\measurebox}
\usepackage{multirow}

\usepackage[breaklinks=true,bookmarks=false]{hyperref}

\iccvfinalcopy 

\ificcvfinal\fi
\begin{document}

\title{Down-Scaling with Learned Kernels in Multi-Scale Deep Neural Networks\\ for Non-Uniform Single Image Deblurring}

\author{Dongwon Park$^1$ \qquad Jisoo Kim$^1$ \qquad Se Young Chun\\
Ulsan National Institute of Science and Technology (UNIST)\\
UNIST-gil 50, Ulsan, Republic of Korea\\
{\tt\small \{dong1,rlawltn1053,sychun\}@unist.ac.kr}, {\small $^1$equal contribution.}
}

\maketitle
\thispagestyle{empty}


\begin{abstract}
Multi-scale approach has been used for blind image / video deblurring problems to yield excellent performance for both conventional and recent deep-learning-based state-of-the-art methods. Bicubic down-sampling is a typical choice for multi-scale approach to reduce spatial dimension after filtering with a fixed kernel. However, this fixed kernel may be sub-optimal since it may destroy important information for reliable deblurring such as strong edges. We propose convolutional neural network (CNN)-based down-scale methods for multi-scale deep-learning-based non-uniform single image deblurring. We argue that our CNN-based down-scaling effectively reduces the spatial dimension of the original image, while learned kernels with multiple channels may well-preserve necessary details for deblurring tasks. For each scale, we adopt to use RCAN (Residual Channel Attention Networks) as a backbone network to further improve performance. Our proposed method yielded state-of-the-art performance on GoPro dataset by large margin. Our proposed method was able to achieve 2.59dB higher PSNR than the current state-of-the-art method by Tao. Our proposed CNN-based down-scaling was the key factor for this excellent performance since the performance of our network without it was decreased by 1.98dB. The same networks trained with GoPro set were also evaluated on large-scale Su dataset and our proposed method yielded 1.15dB better PSNR than the Tao's method. Qualitative comparisons on Lai dataset also confirmed the superior performance of our proposed method over other state-of-the-art methods.
\end{abstract}

\section{Introduction}

\begin{figure}[!t]
	\centering
	\includegraphics[width=1.0\linewidth]{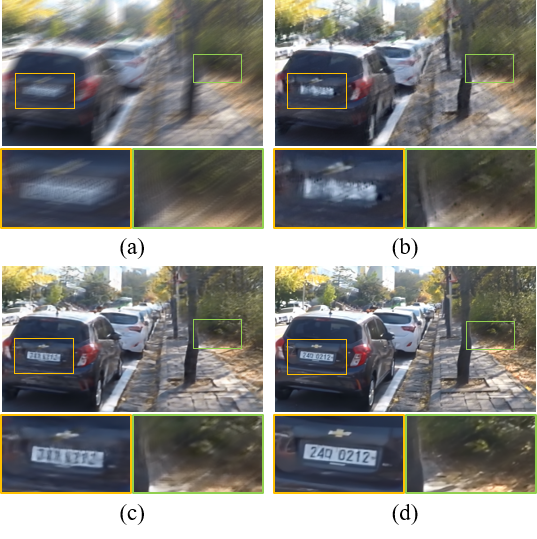}
		\vskip -0.1in
	\caption{
		A deblurring example. (a) Input blurred image, (b) Results of Tao~\cite{Tao:2018vg}, (c) Results of Nah~\cite{Nah:2017bx}, and our results. A car license plate is clearly recovered by our proposed method while other methods can not.}
	\label{fig:relatedwork_p}
	\vskip -0.1in
\end{figure}

Images and videos are often blurred by various reasons: unwanted camera shake, moving objects, depth variation in the scene. A typical forward model to generate a blurred image ${\bf y} \in \mathbb{R}^{N}$ with $N$ number of pixels is
\begin{equation}
    \bf{y} = \bf{G} \bf{x} + \bf{n}
    \label{eq:deblurmodel}
\end{equation}
where ${\bf G} \in \mathbb{R}^{N \times N}$ is a unknown large sparse matrix whose rows contain local blur kernels, ${\bf x}\in \mathbb{R}^{N}$ is a unknown latent ground truth image that one would like to estimate, and ${\bf n}\in \mathbb{R}^{N}$ is noise.
The goal of blind single image deblurring is to recover the values of ${\bf G}$ and ${\bf x}$ for the given ${\bf y}$. Thus, it is a challenging inverse problem that is severely ill-posed.

There have been much effort to tackle blind single image / video deblurring. One is to simplify (\ref{eq:deblurmodel}) by assuming uniform blur and to recover both the latent ground truth image $\bf{x}$ and the blur kernel matrix $\bf{G}$ by solving a tractable optimization problem based on (\ref{eq:deblurmodel})~\cite{Fergus:2006ch,Shan:2008ef,Cho:2009dr,Xu:2010cr}. However, uniform blur is often not accurate enough to approximate the actual blur due to, for example, camera rotation and movements outside the sensor plane or moving objects in the scene with different distances from the camera. Thus, there have been much research on removing non-uniform blur by extending the degree of freedom of the blur model from uniform blur to non-uniform blur in a limited way 
compared to the dense matrix $\bf{G}$ in (\ref{eq:deblurmodel})~\cite{Harmeling:2010we,Gupta:2010bv,Whyte:2010ct,Hirsch:2011fw,Xu:2013tl,Pan:2016ve}. Other 
non-uniform blur models have been investigated such as additional segmentations within which simple blur models were used~\cite{CouzinieDevy:2013bf,Kim:2013dg} or motion estimation based deblurs~\cite{Kim:2014gn,Kim:2015bw}.

Recently, deep-learning-based approaches for single image / video blind deblurring have been proposed with excellent quantitative results and with fast computation time. There are largely two different ways of using deep neural networks for deblurring. One is to use neural networks to explicitly estimate non-uniform blurs $\bf{G}$~\cite{Sun:2015je,Chakrabarti:2016bj,Schuler:2016fk,Bahat:2017ez} and the other is to use networks to directly estimate the original sharp image ${\bf x}$ without estimating blurs~\cite{Xu:2014wh,Kim:2017hk,Wieschollek:2017iy,Su:2017bk,Nah:2017bx,Tao:2018vg}. Current state-of-the-art methods are the work of Nah \textit{et al.}~\cite{Nah:2017bx} and the work of Tao \textit{et al.}~\cite{Tao:2018vg} that are estimating the original sharp image directly from the given blurred image ${\bf y}$ and are using so-called multi-scale (coarse-to-fine) approaches with down-scaled image(s).

Multi-scale approaches are popular in many low-level computer vision tasks such as depth map prediction~\cite{Eigen:2014vq}, surface normal / semantic label predictions~\cite{Eigen:2015bs}, image / video deblurring~\cite{Su:2017bk,Nah:2017bx,Tao:2018vg} as well as high-level computer vision tasks such as image generation~\cite{Denton:2015to}, video frame prediction~\cite{Mathieu:2016wv}. There are two types of generating down-scaled image (or information): 1) down-sampling after filtering with a fixed kernel such as Gaussian or bicubic so that local information will be encoded with reduced spatial dimension~\cite{Fergus:2006ch,Cho:2009dr,Denton:2015to,Mathieu:2016wv,Nah:2017bx,Tao:2018vg} and 2) down-scaling with deep neural networks so that global information will be encoded with further reduced spatial dimension~\cite{Eigen:2014vq,Eigen:2015bs,Su:2017bk}. The former seems to lose much high-frequency information during down-sampling process in a sub-optimal way for image deblurring considering the fact that strong edge information is important for reliable deblurring~\cite{Cho:2009dr,Xu:2010cr}. The latter intended to encode all global information with greatly reduced spatial dimension (e.g., up to $\times8$~\cite{Su:2017bk}) and increased number of channels (e.g., up to 512 channels~\cite{Su:2017bk}). Thus, there have been no method to have the advantages of both approaches.

In this article, we investigated multi-scale approach to single image deblurring problems. First of all, we propose a novel convolutional neural network (CNN)-based down-scaling method that is in between conventional down-sampling methods with fixed kernel and recent deep neural network based methods with global information encoding. We argue that our CNN-based down-scaling allows to perform spatial dimension reduction with learned kernels to encode local information such as strong edges and keeps the number of channels at each scale so that too much global information is not encoded locally after down-scaling. With our CNN-based down-scaling modules, we propose a deep multi-scale single image deblurring network based on RCAN (Residual Channel Attention Networks)~\cite{Zhang:2018dw} that is a state-of-the-art method for super resolution as a backbone network. Unlike the work of Nah \textit{et al.}~\cite{Nah:2017bx}, our proposed method used our down-scaling with learned kernels and RCAN, 
employed sub-optimal modular training approach instead of end-to-end training for deeper network with limited hardware resource, and removed the deblur network to process the original blurred image in the multi-scale network without compromising performance.

\section{Related Works}

Conventional approaches to blind single image / video deblurring usually require to explicitly estimate blur kernels. There have been several works on estimating uniform blurs using optimization algorithm with coarse-to-fine multi-scale approach~\cite{Fergus:2006ch}, using a model of the spatial randomness of noise and a local smoothness prior~\cite{Shan:2008ef}, exploiting blurred strong edges to reliably estimate blur kernel~\cite{Cho:2009dr}, and developing a metric to measure the usefullness of image edges for blur kernel estimation~\cite{Xu:2010cr}.

There have also been many works on predicting non-uniform blurs assuming spatially linear blur~\cite{Harmeling:2010we}, simplified camera motion (from 6D to 3D)~\cite{Gupta:2010bv}, parametrized geometric model in terms of camera rotation velocity during exposure~\cite{Whyte:2010ct}, filter flow framework based blur model~\cite{Hirsch:2011fw}, L0 sparse expression for blurs~\cite{Xu:2013tl}, and dark channel prior~\cite{Pan:2016ve}. There was also an attempt to exploit multiple images from videos assuming spatially varying blur~\cite{Li:2010gy}. There have also been some works to utilize segmentation information by assuming uniform blur on each segmentation area~\cite{CouzinieDevy:2013bf} and to segment motion blur using convex optimization~\cite{Kim:2013dg}, to simplify motion model as local linear without segmentation using coarse-to-fine approach~\cite{Kim:2014gn}, and to use bidirectional optical flows for video deblurring~\cite{Kim:2015bw}.

Since the advent of deep learning~\cite{LeCun:2015dt}, many blind single image / video deblurring works employed deep neural networks for estimating blur kernels and/or original sharp images from given blurred input images. There are several works to predict non-uniform blur kernels explicitly: predicting the probabilistic distribution of motion blur at the patch level~\cite{Sun:2015je}, estimating the complex Fourier coefficients of a deconvolution filter~\cite{Chakrabarti:2016bj}, performing blur kernel estimation by division in Fourier space from extracted deep features~\cite{Schuler:2016fk}, and analyzing the spectral content of blurry image patches by reblurring them~\cite{Bahat:2017ez}. 

\begin{figure*}[!t]
	\centering
	\includegraphics[width=1.0\linewidth]{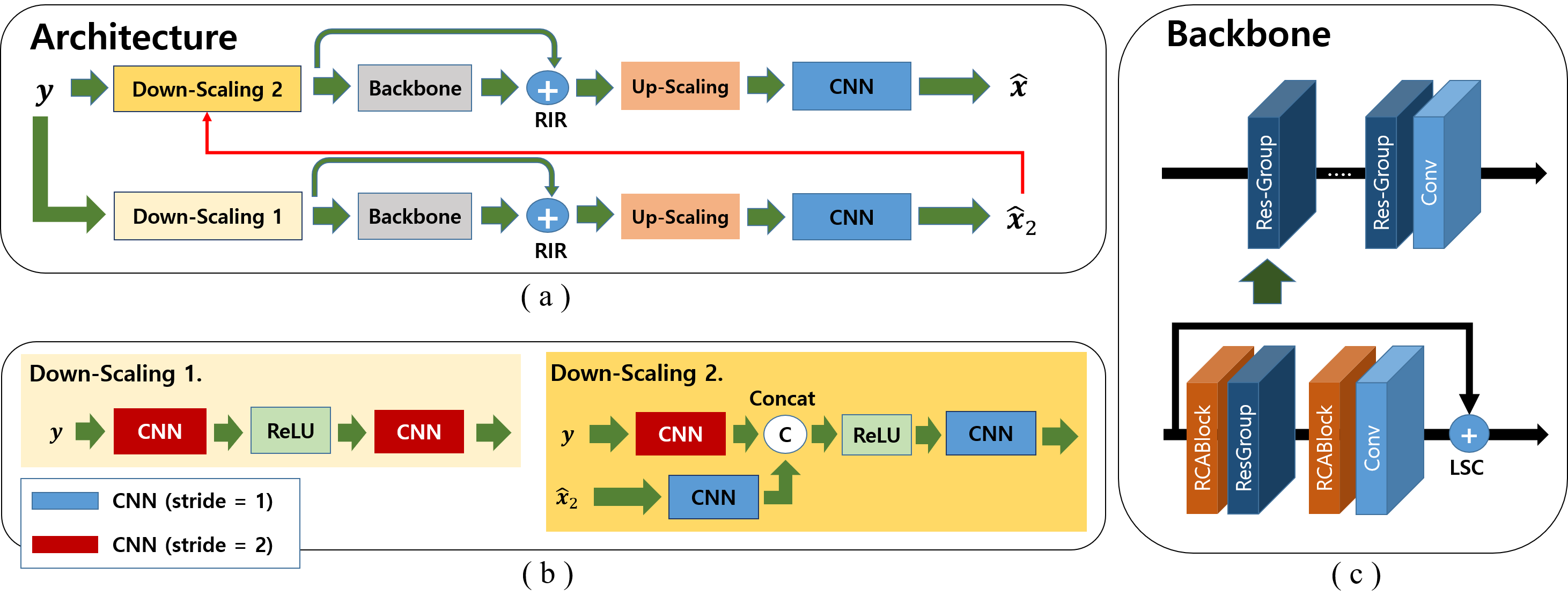}
	\vskip -0.1in
	\caption{(a) Our proposed multi-scale architecture. Note that there is no network to directly process images at the original scale. (b) Our proposed down-scaling modules with learned kernels. (c) Our adopted RCAN backbone network~\cite{Zhang:2018dw}.}
	\label{fig:Model Architecture}
		\vskip -0.1in
\end{figure*}

There are also many works to directly estimate the original sharp image from the given blurred input image without explicitly estimating non-uniform blur kernels. For video blind deblurring, there have been some works to exploit temporal information:
blending temporal information in spatio-temporal recurrent network for online video deblurring~\cite{Kim:2017hk}, taking temporal information into account with recurrent deblur network consisting of several deblur blocks~\cite{Wieschollek:2017iy}, and developing an encoder-decoder network with the input of multiple video frames to accumulate information across frames~\cite{Su:2017bk}.
There are a few works 
for blind single image deblurring without temporal information.
Xu \textit{et al.} proposed a direct estimation method of the original sharp image based on conventional optimization to approximate 
deconvolution by a series of convolution steps using deep neural networks~\cite{Xu:2014wh}. Later, Nah~\textit{et al.} proposed a multi-scale network architecture with Gaussian pyramid and multi-scale loss functions~\cite{Nah:2017bx} and Tao~\textit{et al.} proposed convolution long short-term memory (LSTM)-based multi-scale deep neural network for single image deblurring~\cite{Tao:2018vg}.




Multi-scale approaches for single image / video deblurring have two different types of down-scaling. One is a simple filtering \& down-sampling operation so that local information will be encoded with reduced spatial dimension~\cite{Fergus:2006ch,Cho:2009dr,Nah:2017bx,Tao:2018vg}. The other is a deep neural network based global information encoding of multiple video frames (e.g., 5)
with much further reduced spatial dimension (up to $\times8$) and increased channels (up to 512)~\cite{Su:2017bk}. 
There has been no work on learning based down-scaling to encode local information as an extension of Gaussian / bicubic down-sampling in multi-scale single image deblurring.

\section{Method}

\subsection{Proposed Network Architecture}

Figure~\ref{fig:Model Architecture} illustrates the overall architecture of our proposed deep neural network model for single image deblurring. While conventional multi-scale approaches~\cite{Nah:2017bx,Tao:2018vg}  are using Gaussian / bicubic downsampling with a fixed kernel to reduce the spatial dimension of the input blurred image, we propose to use a CNN-based downscaling with learned kernels to preserve necessary high-frequency information for deblurring. Our network firstly down-scales the input color image with the size of $W\times H\times3$ to the feature maps with the size of $W/4 \times H/4 \times 64$ using the ``Down-Scaling 1'' module in Figure~\ref{fig:Model Architecture} (b). Then, these feature maps are fed into the backbone network as well as residual in residual (RIR) skip connection to yield initial deblurred feature maps in the spatial dimension of $W/4 \times H/4$. Then, up-scaling and CNN will result in the intermediate deblurred image estimate with the size of $W/2 \times H/2 \times 3$. This result is combined with the down-scaled blurred image in the ``Down-Scaling 2'' module in Figure~\ref{fig:Model Architecture} (b) for the deblurring at the scale of $W/2 \times H/2$ using another backbone network, RIR, and up-scaling process to yield the final deblurred output image with the size of $W \times H$. Unlike other multi-scale approaches for single image deblurring~\cite{Nah:2017bx,Tao:2018vg}, our proposed network does not process the original blurred image at the finest scale so that less parameters are required to be used and computation complexity is significantly reduced. 
We empirically confirmed that adding the module to directly process the original blurred image did not help to improve the overall performance.


\subsection{Proposed Down-Scaling with Learned Kernels}

Gaussian or Bicubic down-sampling is a simple, effective method to reduce the spatial dimension of an image with a fixed kernel. For the input image with the size of $W \times H \times 3$, bicubic $\times 2$ down-sampling yields $W/2 \times H/2 \times 3$ image. However, due to the fixed kernel, some important high-frequency information for deblurring may be removed to avoid aliasing. For example, it is well-known that strong edges help to reliably estimate blur kernels~\cite{Cho:2009dr,Xu:2010cr}, thus it may be beneficial to preserve some of these high-frequency details even in down-scaled information for deblurring.

We propose CNN-based down-scaling modules instead of using conventional down-sampling. For $\times 4$ down-scaling, our proposed module consists of convolution layer - ReLU - convolution layer where each convolution layer has 64 channels and stride 2 as illustrated in Figure~\ref{fig:Model Architecture} (b). 
For $\times 2$ down-scaling in multi-scale architecture, our proposed architecture consists of convolution layer with stride 2 - concatenation with the output from the lower scale deblurring network - ReLU - convolution later with stride 1 as illustrated in Figure~\ref{fig:Model Architecture} (b). All convolution layers have $3\times 3$ filters that have similar sizes as bicubic kernel.

\begin{figure}[!t]
	\centering
	\includegraphics[width=1.0\linewidth]{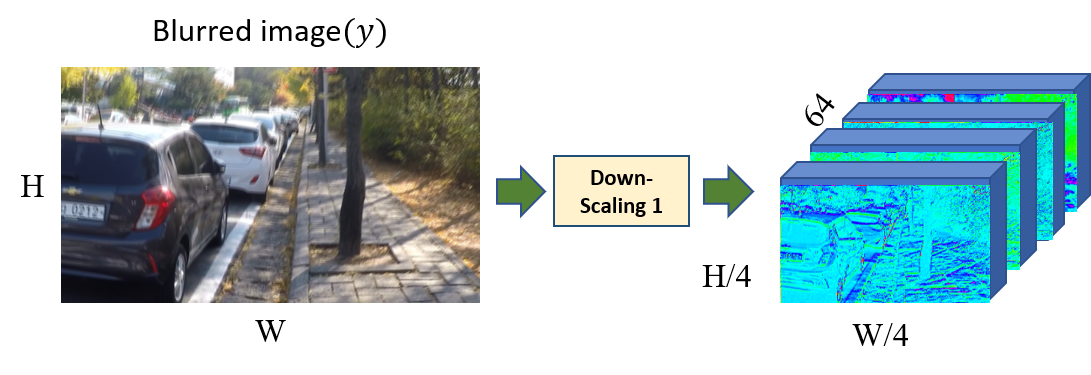}
	\vskip -0.1in
	\caption{An example of our CNN-based down-scaling.}
	\label{fig:CNN_down_scaling}
		\vskip -0.1in
\end{figure}

Unlike conventional down-sampling after anti-aliasing filtering with a deterministic kernel, our proposed CNN-based down-scaling has learned kernels with 64 channels and spatial dimension reduction using stride 2. Thus, our proposed method seems to keep the effect of spatial dimension reduction for multi-scale approach while to be more optimal than anti-aliasing filtering (essentially low-pass filtering). Figure~\ref{fig:CNN_down_scaling} illustrates one example of our CNN-based down-scaling. Our proposed method seems to preserve some high-frequency information (e.g., strong edges) that can be essential for reliable deblurring~\cite{Cho:2009dr,Xu:2010cr}.
Our proposed down-scaling with learned kernels is also different from the work of~\cite{Su:2017bk} and others that are using a deep neural network based global information encoding with further reduced spatial dimension and increased channels. Note that the work of~\cite{Su:2017bk} and others do not yield any intermediate deblurred image estimate at low scales, have much larger receptive field size than bicubic kernels so that less local information is preserved, and 
have increased channels for global information encoding so that it can not be seen as an extension of conventional bicubic down-sampling.


\subsection{Residual Channel Attention Network}

A backbone network architecture is often shared among single image deblurring and single image super resolution. For example, the work of Nah~\cite{Nah:2017bx} for deblurring and the work of Lim (EDSR)~\cite{lim2017enhanced} for super resolution utilized the same residual block as their backbone networks.
Recently, residual channel attention network (RCAN) was proposed for super resolution and yielded state-of-the-art results~\cite{Zhang:2018dw}.
We adopt RCAN as our backbone network for single image deblurring.
RCAN contains a small number of network channels, but had a large receptive field due to deep architecture. RCAN consists of a number of residual groups with long skip connections (LSC) and each residual group consists of multiple residual channel attention blocks with short skip connections (SSC). RCAN with large receptive field was able to be well-trained due to SSC as well as LSC.


\begin{figure}[!t]
	\centering
	\includegraphics[width=1.0\linewidth]{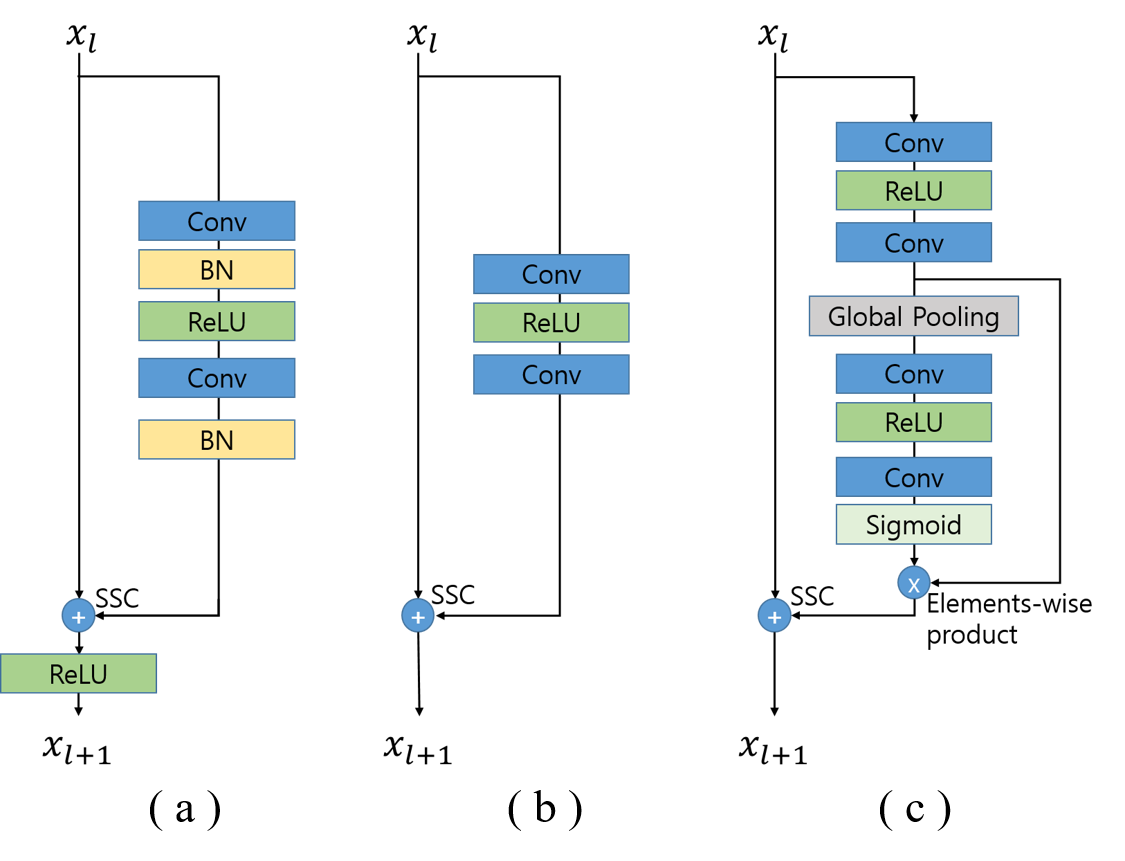}
	\vskip -0.1in
	\caption{(a) Residual blocks in ResNet~\cite{he2016deep}. (b) Residual blocks in EDSR~\cite{lim2017enhanced}, the work of Nah~\cite{Nah:2017bx}. (c) Our adopted residual channel attention blocks~\cite{Zhang:2018dw}.}
	\label{fig:resblocks}
		\vskip -0.1in
\end{figure}

Figure~\ref{fig:resblocks} illustrates the differences among the original residual blocks of ResNet~\cite{he2016deep} 
for high-level computer vision task such as image classification, a modified residual blocks in ESDR~\cite{lim2017enhanced} and the work of Nah~\cite{Nah:2017bx} for low-level computer vision tasks such as super resolution and image deblurring, and the residual channel attention blocks for super resolution~\cite{Zhang:2018dw}. Note that there is no batch normalization layer for super resolution and image deblurring. RCAN additionally combined residual blocks with channel attention mechanism that generated global context information through global average pooling. This information combined with the output of residual block through element-wise product to yield the final output as in Figure~\ref{fig:resblocks} (c). 
Recent image super resolution methods~\cite{lim2017enhanced,ledig2017photo} including RCAN are using pixel shuffle
for up-scaling due to low computation complexity and high performance. We also adopt pixel shuffle for our up-scaling modules.




 
\subsection{Loss Function}

Recently, a mix loss was proposed to combine a conventional L1 loss and a multi-scale SSIM (structural similarity) loss to yield shaper images than L1 loss only~\cite{zhao2017loss}. We used this mix loss for training both our coarse-scale sub-network as well as our fine-scale sub-network as follows:
\begin{equation}
    L_{tot} (\theta_c, \theta_f) = \lambda_{L1} L_{L1}(\theta_c, \theta_f) + \lambda_{SS} L_{SS}(\theta_c, \theta_f)
    \label{eq:loss}
\end{equation}
where $\theta_c, \theta_f$ are deep neural network parameters for the coarse-scale sub-network, the fine-scale sub-network, respectively.
The L1 loss is defined as
\begin{equation}
    L_{L1}(\theta_c, \theta_f) = \rVert I^{GT} - I^{out} \rVert_1 \nonumber
\end{equation}
where  $I^{GT}$ is a ground truth image and
\[
I^{out} = h^f (I^{in}, h^c (I^{in}; \theta_c); \theta_f)
\]
where $I^{in}$ is an input blurred image and $h^c$, $h^f$ are coarse-scale, fine-scale deep neural networks, respectively.
The multi-scale SSIM loss is defined as
\begin{equation}
    L_{SS}(\theta_c, \theta_f) = 1 - \sum_{m = 0}^2 w_m S(D_{ap}^m(I^{out}),D_{ap}^m(I^{GT})) \nonumber
\end{equation}
where $D^m_{ap}$ is a down-scale operator with $\times2^m$ using average pooling. 
The following parameters are selected: $\lambda_{SS}$=0.78, $\lambda_{L1}$ = 0.22, $w_1$ = 0.448, $w_2$ = 0.353, $w_3$ = 0.199.

\subsection{Modular Training Approach}

Previous multi-scale approaches for single image deblurring trained their networks in an end-to-end manner~\cite{Tao:2018vg,Nah:2017bx}. For our loss function, an end-to-end training should minimize (\ref{eq:loss}) in terms of both $\theta_c, \theta_f$ for two sub-networks.
Even though end-to-end training is optimal for estimating all network parameters to yield good performance, it often limits the size of a network
such as depth that is also important for good performance
due to hardware limitation (e.g., GPU memory). Thus, there is a trade-off between end-to-end optimization and the size of a network. We chose to have an effectively deeper network, rather than to perform an end-to-end training on a shallower network. 

We propose a modular training approach for multi-scale architecture. We first trained the sub-network $h^c (I^{in}; \theta_c)$
with the ground truth of $\times2$ bicubic down-sampled image $I^{\times 2,GT}$ by minimizing the following sub-loss function:
\begin{equation}
    L_{sub} (\theta_c) = \lambda_{L1} L^{c}_{L1}(\theta_c) + \lambda_{SS} L^{c}_{SS}(\theta_c)
    \label{eq:subloss}
\end{equation}
where 
\(
L^{c}_{L1}(\theta_c) = \rVert I^{\times 2,GT} - h^c (I^{in}; \theta_c) \rVert_1,
\)
\[
L^{c}_{SS}(\theta_c) = 1 - \sum_{m = 1}^3 w_m S(D_{ap}^m(I^{c,out}),D_{ap}^m(I^{\times2,GT})),
\]
and the intermediate output of our multi-scale approach is
\[
I^{c,out} = h^c (I^{in}; \theta_c).
\]
This sub-optimization will result in the estimate $\hat{\theta}_c$ for (\ref{eq:subloss}).

Then, we finally trained the top network $h^f$ in Figure~\ref{fig:Model Architecture} (a) with the original ground truth image as well as the output of the bottom sub-network $h^c$ in Figure~\ref{fig:Model Architecture} (a) by minimizing (\ref{eq:loss}) with the fixed $\theta_c$ or $L_{tot} (\hat{\theta}_c, \theta_f)$.
Thus, our final trained network is the optimal fine-scale sub-network (the top network in Figure~\ref{fig:Model Architecture} (a)) for the fixed, sub-optimal coarse-scale sub-network (the bottom network in Figure~\ref{fig:Model Architecture} (a)). Due to our modular approach, we were able to train the network that is effectively 2 times deeper than the original RCAN~\cite{Zhang:2018dw}. 


\section{Experimental Setup}

\subsection{Datasets}

GoPro dataset~\cite{Nah:2017bx} for deblurring was used to train our proposed model. This dataset consists of 3,214 blurred images with the size of 1,280$\times$720 that are divided into 2,103 training images and 1,111 test images. We also used K{\"o}hler dataset~\cite{Kohler:2012dj} that contains 48 blurred images with the size of 800$\times$800 for testing. We further evaluated our proposed methods and other state-of-the-art methods on Su dataset~\cite{Su:2017bk}. Su dataset consists of 71 videos (6,708 images) with the size of 1,920$\times$1080 or 1,280$\times$720
from multiple devices including iPhone 6s, GoPro Hero 4, and Canon 7D. 
We used all videos for test.
Lastly, we performed qualitative performance comparisons on Lai dataset~\cite{Lai:2016tg} whose image sizes are varying within 351-1,024$\times$502-1,024.


\subsection{Training Details}

We used PyTorch~\cite{paszke2017automatic} for all our implementations and MATLAB for all PSNR/SSIM evaluation.
NVIDIA Titan V GPU was used for training and testing. Adam optimizer~\cite{kinga2015method} was used with learning rate$=0.5\cdot10^{-5}$, $\beta_1=0.9$, $\beta_2=0.999$, and $\epsilon=10^{-8}$. Total epoch number was 1,000 with reducing learning rate by half every 300 epochs. No pre-training was performed for RCAN backbone network.
Patch based training was performed with the sizes of 192$\times$192 at $\times 4$ scale and 96$\times$96 at $\times 2$ scale with data augmentation using random crop, horizontal flip, and 90 rotation.



\section{Experimental Results}

\subsection{Down-Scaling in Single-Scale Architecture}

\begin{table}[!b]
\caption{Effects of down-scaling at different scales.}
\label{tbl:Down_scale_result}
	\vskip -0.1in
\centering
\begin{tabular}{|c|c|c|r|}
\hline
Measure        & PSNR  & SSIM   & Time   \\ \hline
$\times1\to\times1\to\times1$      & 27.44 & 0.8984 & 21.2 s \\
$\times1\to\times2\to\times1$ & 29.37 & 0.9275 & 4.0 s  \\
$\times1\to\times4\to\times1$ & 31.40 & 0.9501 & 1.0 s  \\ \hline
\end{tabular}
\end{table}

\begin{figure}[!t]
	\centering
	\includegraphics[width=1.0\linewidth]{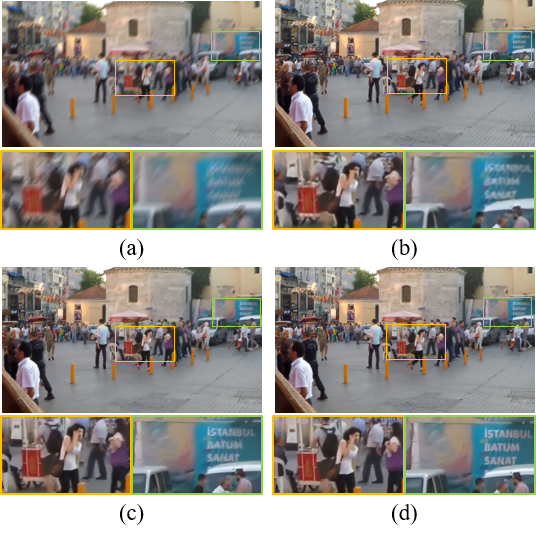}
	\vskip -0.1in
	\caption{An example for the effects of down-scaling at different scales. (a) input blurred image. (b) $\times1\to\times1\to\times1$. (c) $\times1\to\times2\to\times1$. (d) $\times1\to\times4\to\times1$.}
	\label{fig:Down-scale}
		\vskip -0.1in
\end{figure}

First of all, we investigated the effect of our down-scaling approach for different scale levels in single-scale architecture. 
We took out the sub-network $h^c (I^{in}; \theta_c)$ or the bottom network in Figure~\ref{fig:Model Architecture} (a) and modified to yield an image with the original size instead of the $\times2$ down-scaled size. Three different configurations are as follows:
1) $\times1\to\times1\to\times1$ changed stride 2 to 1 in CNNs with no up-scaling, 
2) $\times1\to\times2\to\times1$ changed stride 2 to 1 in the second CNN, and
3) $\times1\to\times4\to\times1$ changed up-scaling to be $\times4$. Table~\ref{tbl:Down_scale_result} summarizes the quantitative results on GoPro test set.
Surprisingly, more down-scaling yielded better PSNR / SSIM and faster computation time.
There are a few explanations for these results such as well-preserved high-frequency details even up to $\times 4$ scale, much larger receptive field for $\times1\to\times4\to\times1$, and reduced blur effect in lower spatial dimensions.
Figure~\ref{fig:Down-scale} also shows qualitative results for this investigation, clearly visualizing superior results of $\times1\to\times4\to\times1$ over others.



\subsection{Ablation Study}

Compared to Nah~\cite{Nah:2017bx}, 
our proposed method contained our CNN-based down-scaling with learned kernels, new backbone network using RCAN~\cite{Zhang:2018dw} and no $\times1\to\times1\to\times1$ path in the multi-scale architecture. 
In order to show the contribution of each component, we performed ablation study:
RCAN vs. EDSR~\cite{lim2017enhanced} for 
backbone network,
CNN-based down-scaling vs. bicubic down-scaling, 
and removing $\times1\to\times1\to\times1$ vs. keeping $\times1\to\times1\to\times1$.

\begin{table}[!b]
\caption{Ablation study results on GoPro dataset}
\label{tbl:Self-Evaluation}
	\vskip -0.1in
\centering
\begin{tabular}{|l|l|l|l|l|r|l|}
\hline
Network & C & $1$ & PSNR  & SSIM   & Time & Param\\ \hline
(a) EDSR     & $\times$               & $\times$          & 29.94 & 0.934 & 2.8s & 80M \\
(b) RCAN     & $\times$               & $\times$          & 30.87 & 0.945 & 8.2s & 32M \\
(c) RCAN     & $\times$               & $\bigcirc$          & 31.23 & 0.948 & 28.6s & 48M \\
(d) RCAN     & $\bigcirc$          & $\times$          & 32.85 & 0.962 & 3.4s & 32M \\
(e) RCAN     & $\bigcirc$          & $\bigcirc$          & 32.88 & 0.962 & 14.6s & 48M \\\hline
\end{tabular}
\end{table}

Table~\ref{tbl:Self-Evaluation} presents PSNR, SSIM, computation time (Time), and parameters size (Param) for different components such as backbone network (Network), CNN-based down-scaling (denoted by C), and $\times 1$-scale network (denoted by 1).
Firstly, RCAN (b) 
yielded substantially better PSNR and SSIM over a modified EDSR (a) 
that was also used in the work of Nah \textit{et al.}~\cite{Nah:2017bx}. Note that EDSR requires 80 million (M) parameters while RCAN does substantially smaller parameters (32M), but EDSR had faster computation time than RCAN possibly due to shallower depth that can utilize parallel GPU computing more efficiently.



Secondly, our proposed CNN-based down-scaling yielded significantly better PSNR (1.98dB up) and better SSIM (0.017 up) than bicubic down-sampling as in Table~\ref{tbl:Self-Evaluation} (d) and (b). Thus, our CNN-based down-scaling is the key factor to improve the overall performance of our proposed single image deblurring among all new components.


Lastly, we investigated the role of $\times1$ path in our proposed method. Table~\ref{tbl:Self-Evaluation} (e) shows that this original scale path did not improve quantitative results with substantially increased computation time and parameter size when using our CNN-based down-scaling.
However, as shown in Table~\ref{tbl:Self-Evaluation} (c), it was important to use $\times1$ path to improve the quality of deblurred images substantially when uing conventional down-sampling.
Thus, we can argue that our CNN-based down-scaling well-preserved necessary high-frequency details at lower scales compared to bicubic down-sampling so that the original input image at the fine scale did not help much to improve the quality of deblurred images.

\subsection{Benchmark Results}

\begin{table}[!b]
\caption{Quantitative results on GoPro and K{\"o}hler datasets. Ours is our network trained with GoPro training dataset.} 
\label{table:GoPro}
	\vskip -0.1in
\centering
\begin{tabular}{|c|c|c|c|c|r|}
\hline
\multirow{2}{*}{ } & \multicolumn{2}{c|}{GoPro}                    & \multicolumn{2}{c|}{K{\"o}hler}                   & \multirow{2}{*}{Time} \\ \cline{2-5}
                         & PSNR                  & SSIM                  & PSNR                  & MSSIM               &                       \\ \hline
Xu~\cite{Xu:2013tl}                       & 25.10                  & 0.890                 & 27.47                 & 0.811                 & 13.41s               \\
Kim~\cite{Kim:2014gn}                      & 23.64                 & 0.824                 & 24.68                 & 0.794                 & 1h                \\
Sun~\cite{Sun:2015je}                      & 24.64                 & 0.843                 & 25.22                 & 0.774                 & 20m                \\
Gong~\cite{gong2017motion}                     & 27.19                 & 0.908                 & 26.59                 & 0.808                 & -                     \\
Ram.~\cite{Ramakrishnan_2017_ICCV}                      & 28.94                 & 0.922                 & 27.08                 & 0.812                 & -                     \\
Nah~\cite{Nah:2017bx}                      & 29.08                 & 0.914                 & 26.48                 & 0.808                 & 3.09s                \\
Kupyn~\cite{kupyn2018deblurgan}                    & 28.70                 & 0.958                 & 26.10                 & 0.816                 & 0.85s                \\
Tao~\cite{Tao:2018vg}                      & 30.26                 & 0.934                 & 26.75                 & 0.837                & 1.87s                \\ \hline
Ours                     & 32.85                 & 0.962                 & 26.08                 & 0.810                & 2.60s                \\\hline
\end{tabular}
\end{table}
%
\begin{table}[!b]
\caption{Quantitative results on Su dataset for the networks trained with GoPro training dataset.}
\label{table:Su}
	\vskip -0.1in
\centering
\begin{tabular}{|c|c|c|c|} 
\hline
     & Nah~\cite{Nah:2017bx}   & Tao~\cite{Tao:2018vg}   & Ours  \\ \hline
     PSNR &  30.34 & 30.93 & 32.08 \\
    SSIM &   0.917 & 0.931 & 0.940 \\ \hline
\end{tabular}
\end{table}

\begin{figure*}[!h]
	\centering
	\includegraphics[width=1.0\linewidth]{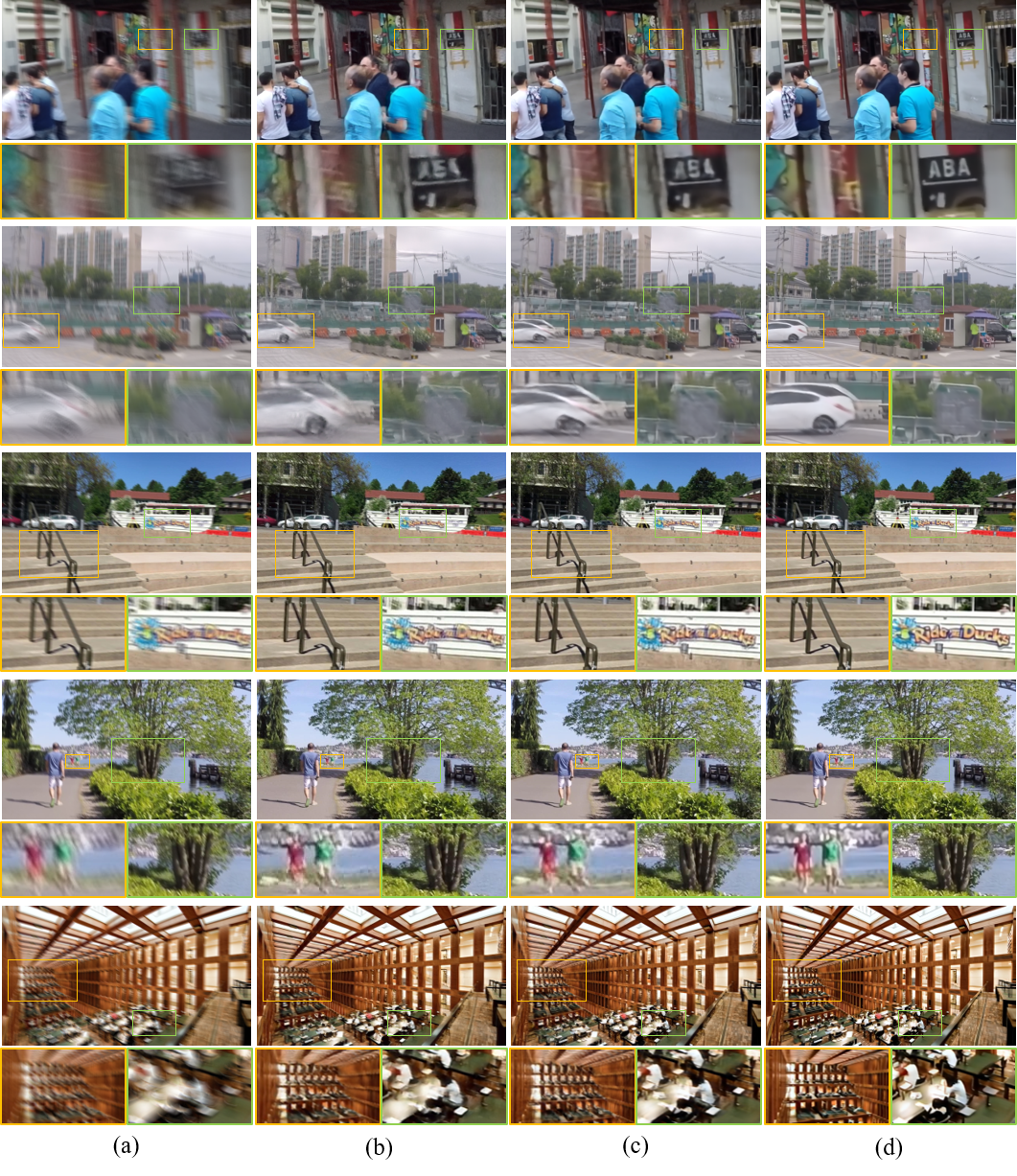}
\vskip -0.1in
	\caption{(a) Input blurred images. (b) Results of Tao~\textit{et al.}~\cite{Tao:2018vg}. (c) Results of Nah~\textit{et al.}~\cite{Nah:2017bx}, (d) Results of our proposed method. 
	The results on the 1st-2nd are from GoPro dataset~\cite{Nah:2017bx},
	those on the 3rd-4th rows are from Su dataset~\cite{Su:2017bk},
	and those on the 5th row are from Lai dataset~\cite{Lai:2016tg}.
	Our proposed method deblurred fine details clearly compared to other state-of-the-art methods 
	on all three datasets.}
	\label{fig:model}
\end{figure*}


We performed comparison studies on a few open datasets such as GoPro, K{\"o}hler, and Su. 
Tables~\ref{table:GoPro} and \ref{table:Su} present quantitative results of our proposed method and other state-of-the-art methods. Our proposed method trained with GoPro training dataset yielded state-of-the-art quantitative results over previous state-of-the-art deblurring methods such as Tao~\cite{Tao:2018vg} in PSNR and Kupyn~\cite{kupyn2018deblurgan} in SSIM on GoPro test dataset (1,111 images). 
Our proposed method yielded 2.59dB better PSNR and 0.028 better SSIM than Tao and also yielded 4.15dB better PSNR and 0.004 better SSIM than Kupyn on GoPro test set. 

Our method trained with GoPro dataset was also evaluated with K{\"o}hler dataset. As shown in Table~\ref{table:GoPro}, our method was 
able to achieve comparable results to the work of Kupyn~\cite{kupyn2018deblurgan}, but lower than Tao~\cite{Tao:2018vg}.
In fact, recent state-of-the-art methods were not able to perform well with K{\"o}hler dataset in terms of PSNR compared to early work of Xu~\cite{Xu:2013tl}.
Note that even though K{\"o}hler dataset contains 48 blurred images, they were actually generated by using 12 different blurs for 4 different images. Thus, the results with this dataset could be severely biased for these 4 images and it may explain rather inconsistent improvements of previous methods between GoPro and K{\"o}hler datasets.

For fair evaluations, 
we further evaluated our method as well as two other state-of-the-art methods by Nah and Tao with all images in Su dataset. 
The codes that were provided by the authors were used.
As shown in Table~\ref{table:Su}, our proposed method yielded significantly better PSNR / SSIM than the work of Nah and the work of Tao. These results are consistent with the results for GoPro test dataset even though Su dataset contains data from other devices.

Figure~\ref{fig:model} shows qualitative comparisons for five different examples from GoPro dataset (1-2 rows), Su dataset (3-4 rows) and Lai dataset (5 row) using our proposed method (d column) as well as the work of Nah (c column) and the work of Tao (b column)
for given input blurred images (a column). Visual comparisons clearly show the superior performance of our proposed method over previous state-of-the-art methods, especially for fine details such as texts, numbers, walking people, and moving cars.


\begin{figure}[!t]
	\centering
	\includegraphics[width=1.0\linewidth]{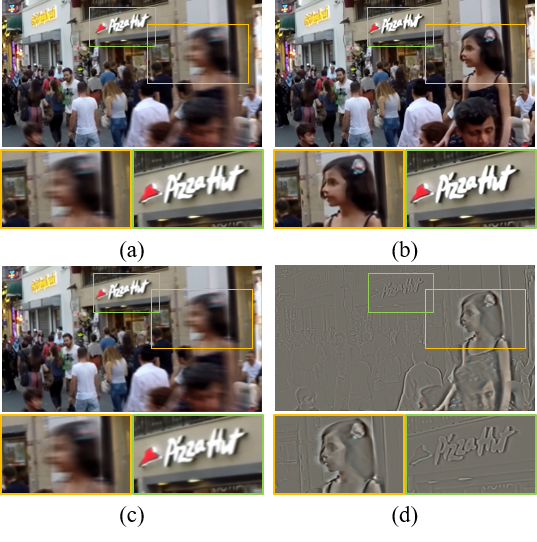}
	\vskip -0.1in
	\caption{Decomposition study. (a) Input blurred image. (b) Output deblurred image. (c) Output without using backbone networks. (d) Output without RIR skip connections.}
	\label{fig:Bicubic_down}
		\vskip -0.1in
\end{figure}

\section{Discussion}

We propose a CNN-based down-scaling method with learned kernels for multi-scale deep neural networks in single image deblurring. Unlike other multi-scale deblurring methods with Gaussian / bicubic down-sampling, our proposed down-scaling seems to preserve necessary high-frequency details such as strong edges for reliable and high performance deblurring even at reduced spatial dimension. Our scheme also allows us to use no network processing the original scale data with negligible performance degradation so that computation complexity is not increased. Our proposed method yielded state-of-the-art performance on large-scale datasets such as GoPro and Su. Our method also yielded visually pleasing deblurring results on Lai dataset compared to other state-of-the-art methods.

Even though our proposed method yielded similar results to other state-of-the-art methods on K{\"o}hler dataset. Most previous methods and our proposed method did not yield consistent performance gain. Methods with excellent performance on GoPro dataset often failed to get similar state-of-the-art results on K{\"o}hler dataset. Even though it seems true that the results on K{\"o}hler dataset could be biased due to small number of images in it, training data must be carefully investigated. For example, GoPro and K{\"o}hler datasets have different image sizes, used different ways of generating blurs, and different scenes and environments.
Thus, we argue that GoPro dataset may not be the best training set for K{\"o}hler test set.
Our observation can also be seen in other literature: Kupyn \textit{et al.} investigated two training datasets (ImageNet based synthetic blur dataset and GoPro dataset) and obtained quite different performance results~\cite{kupyn2018deblurgan}. Thus, there have been much research on blur data generation~\cite{Schuler:2016fk,Kim:2017hk,Nah:2017bx,gong2017motion,kupyn2018deblurgan}.
As an example, we fine-tuned our GoPro-trained network with Su training dataset with several epochs only and it yielded 27.26dB on K{\"o}hler dataset. This shows the importance of training dataset for performance.
Further investigation is needed to address this.

Lastly, Figure~\ref{fig:Bicubic_down} shows one example of the input blurred image, the output deblurred image by our proposed method, the output image without using backbone networks, and the output image without RIR skip connections. This decomposition study illustrates the roles of each component in our method. Even though our training process does not enforce to separate the role of each component, backbone networks estimate blurring information that is  corresponding to the blur kernel estimation in conventional deblurring methods.

 \section*{Acknowledgments}
 
  This work was supported partly by 
  Basic Science Research Program through the National Research Foundation of Korea(NRF) 
  funded by the Ministry of Education(NRF-2017R1D1A1B05035810),
  the Technology Innovation Program or Industrial Strategic Technology Development Program 
  (10077533, Development of robotic manipulation algorithm for grasping/assembling 
  with the machine learning using visual and tactile sensing information) 
  funded by the Ministry of Trade, Industry \& Energy (MOTIE, Korea), and a grant of the Korea Health Technology R\&D Project 
  through the Korea Health Industry Development Institute (KHIDI), 
  funded by the Ministry of Health \& Welfare, Republic of Korea (grant number: HI18C0316).


{\small
\bibliographystyle{ieee}
\bibliography{DSLK_v3}
}

\end{document}